\documentclass[11pt,a4paper]{article}
\usepackage[hyperref]{acl2020}
\usepackage{times}
\usepackage{latexsym}

\usepackage{microtype}
\usepackage{amsmath}
\usepackage{amsfonts}
\usepackage{mathrsfs}
\usepackage{cleveref}
\usepackage[switch]{lineno}
\usepackage{xcolor}
\usepackage{graphicx}
\usepackage{multirow}

\aclfinalcopy

\title{Open-Domain Dialogue Generation Based on Pre-trained Language Models}

\author{Yan Zeng \\
  DIRO, Université de Montréal \\
  \texttt{yan.zeng@umontreal.ca} \\\And
  Jian-Yun Nie \\
  DIRO, Université de Montréal \\
  \texttt{nie@iro.umontreal.ca} \\}

\date{}

\begin{document}
\maketitle
\begin{abstract}
Pre-trained language models have been successfully used in response generation for open-domain dialogue. Four main frameworks have been proposed: (1) Transformer-ED using Transformer encoder and decoder separately for source and target sentences; (2) Transformer-Dec using Transformer decoder for both source and target sentences; (3) Transformer-MLM using Transformer decoder that applies bi-directional attention on the source side and left-to-right attention on the target side with masked language model objective; and (4) Transformer-AR that uses auto-regressive objective instead. In this study, we compare these frameworks on 3 datasets, and our comparison reveals that the best framework uses bidirectional attention on the source side and does not separate encoder and decoder. We also examine model discrepancy, and our experiments confirm that the performance of a model is directly impacted by the underlying discrepancies. We then propose two correction methods to reduce the discrepancies, and both improve the model performance. These results show that discrepancies is an important factor to consider when we use a pre-trained model, and a reduction in discrepancies can lead to improved performance.
\end{abstract}

\section{Introduction} 
General purpose (non goal-oriented) dialogue has been investigated using data-driven sequence-to-sequence (SEQ2SEQ) recurrent neural networks (RNN) \cite{sordoni2015neural, shang2015neural, wen2015semantically, vinyals2015neural}. Recently, fine-tuning pre-trained language models has demonstrated superior performance in the ConvAI2 \cite{dinan2019second} competition. 
Several other studies also successfully exploited  pre-trained language models for dialogue generation. It is intuitive that the rich language knowledge encoded in the model pre-trained on a large amount of raw texts can help dialogue systems to generate more reasonable responses. While one can easily agree on this principle, it is much more difficult to determine how a language model should be adapted for dialogue generation.
Different frameworks have been proposed in the literature:
Transformer-ED (explicit encoder and decoder architecture) \cite{zheng2019pre}, Transformer-Dec (decoder only) \cite{wolf2019transfertransfo,lin2019caire,zhang2019dialogpt}, Transformer decoder that uses bi-directional attention on the source side and left-to-right attention on the target side with Masked Language Model (MLM) objective (Transformer-MLM) \cite{dong2019unified} or with Auto-Regressive (AR) objective (Transformer-AR) \cite{bao2019plato, shuster2019dialogue}. The first two frameworks utilize Generative Pre-Training (GPT) \cite{radford2018improving}, a left-to-right architecture pre-trained with AR. 
The left-to-right generation fashion corresponds well to that of dialogue response generation. Thus, some researchers believe that this framework naturally works well for dialogue response generation \cite{lin2020exploring}. The two latter frameworks use BERT \cite{devlin2018bert}, a bi-directional architecture pre-trained with MLM. BERT has been used widely as the encoder for classification tasks \cite{zhang2019find, zeng2020multidomain}, while some studies \cite{dong2019unified, zeng2020generalized} show that fine-tuning BERT can also achieve state-of-the-art performance for response generation in dialogue. To our knowledge, no study has investigated into the fine-tuning methods to understand why and how a pre-trained language model can be fine-tuned for dialogue generation. The choice of a pre-trained language model and the way to adapt it are still an art.

In this study, we aim at providing an analysis about the utilization of pre-trained language models for dialogue generation. To this end, we re-implement the existing approaches proposed in the literature and run extensive experiments on 3 datasets to compare them.  Our results show that Transformer-ED that separates encoder and decoder does not produce competitive results against others that combine them, and models that use bi-directional attention to encode dialogue history outperforms the one using unidirectional (left-to-right) attention. However, an advantage of using unidirectional attention is generating diverse responses. Additionally, this comparison reveals some important aspects that were neglected in the utilization of pre-trained models, namely, the discrepancies that may occur between pre-training and the fine-tuning processes and between fine-tuning and the generation (inference) process.

The concept of model discrepancy 
has been briefly mentioned in \citet{yang2019xlnet} to mean that the model has been trained in a way, but used in a different way for the task. However, the problem has not been investigated in depth. We believe that model discrepancy is a very important aspect that can bring a significant impact on the final result. Going further in this direction, we define two discrepancies: \textbf{pretrain-finetune discrepancy} which means the differences in architecture and loss function between pre-training and fine-tuning, and \textbf{finetune-generation discrepancy} which means that the way it is used in generation (inference/test) is different from the way it has been trained. For the four frameworks we mentioned for dialogue generation based on pre-trained models, except Transformer-Dec, they all have some pretrain-finetune discrepancies, while only Transformer-MLM has finetune-generation discrepancy because of MLM objective: during training, the model input has random masks, while in generation process, the input does not contain masks (see Figure \ref{Fig:finegen}). We summarize the discrepancies of different models in (Table \ref{tab:frameworks}).

Discrepancies may affect the model performance since models with such discrepancies cannot best exploit the pre-trained language model or the fine-tuned model. The impact of pretrain-finetune discrepancy could be reduced by using a large dataset since a large amount of training data can correct the discrepancies to some extent. This also explains why discrepancy has not attracted much attention -- most of the work on dialogue generation tries to use as much training data as possible. 
However, if the amount of training data is limited, the discrepancy problem may surface. Therefore, we will analyze the performance of the models with both large and small amount of training data in order to make it easier to observe the impact of discrepancies.
In particular, our experiments will show that Transformer-ED and Transformer-AR, which have the largest pretrain-finetune discrepancy, are more impacted than Transformer-MLM and Transformer-Dec by a small amount of data. 

To further confirm that discrepancies are truly the hidden reason of model performance, we propose correction measures to reduce pretrain-finetune discrepancy and finetune-generation discrepancy of Transformer-MLM, in order to see if these measures can avoid the negative impact of discrepancies. These results show that the performance of the model is increased when the discrepancy issues are corrected, confirming that discrepancies are indeed an important factor that influence the effectiveness of a pre-trained model for dialogue generation.
This study is the first investigation to show explicitly the phenomenon of model discrepancy and its impact on performance. It can be seen as an appeal to more investigations on this important problem when we adapt a pre-trained model.

The contributions in this work are summarized as: 
\begin{itemize}
\item We re-implement and compare four major frameworks that utilize pre-trained language models for dialog generation on three public dialogue datasets and we consider two data scales \footnote{We will release our codes later.}. Our experimental results will support our analysis on model's architectural appropriateness.

\item We formally introduce the concept of pretrain-finetune discrepancy and finetune-generation discrepancy when exploit a pre-trained model. We examine the discrepancies of each framework and confirm the impact of discrepancies in experiments.

\item We propose two correction methods to decrease pretrain-finetune discrepancy and finetune-generation discrepancy of Transformer-MLM. Both corrections lead to increased performance of Transformer-MLM, confirming that the correction of discrepancies is an important aspect to consider in utilizing a pre-trained model.
\end{itemize}

\section{Pre-training Based Transformer Frameworks}
\label{sec:related_Trans}
\begin{figure*}[t]
\centering
\includegraphics[height=1.3in]{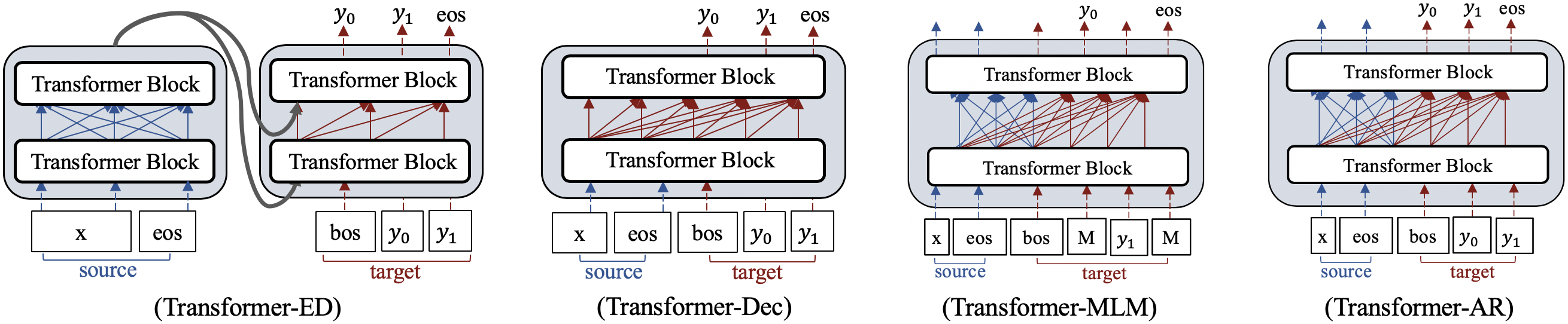}
\caption{Architectures of four pre-training based Transformers for dialogue generation.}
\label{Fig:model}
\end{figure*}

We start with a brief description of the existing approaches to dialogue generation based on pre-trained language models. Figure \ref{Fig:model} and Table \ref{tab:frameworks} provide an overview of them. 

\begin{table*}[t]
\small
\centering
\begin{tabular}{l|llll}
\hline 
\hline
 & \textbf{Transformer-ED} & \textbf{Transformer-Dec} & \textbf{Transformer-MLM} & \textbf{Transformer-AR} \\
\hline 
Pre-trained LM & GPT  & GPT & BERT  & BERT \\

Architecture & {\color{red}encoder-decoder} & decoder-only & decoder-only & decoder-only \\

Source Side Attn. & {\color{red}bi-directional} & left-to-right & bi-directional & bi-directional \\

Target Side Attn. & left-to-right & left-to-right & {\color{red}left-to-right} & {\color{red}left-to-right} \\

Objective & auto-regressive & auto-regressive & Masked-LM & {\color{red}auto-regressive} \\
\hline
\end{tabular}
\caption{\label{tab:frameworks} Key characteristics of the four pre-training based Transformers. Characteristics in {\color{red}red} are inconsistent between pre-training and fine-tuning. 
}
\end{table*}

\subsection{Transformer-ED}
\label{sec:trans_ed}
Transformer-ED is an encoder-decoder architecture as in \citet{vaswani2017attention}, which utilizes Generative Pre-Training (GPT) \cite{radford2018improving} to initialize the encoder and the decoder. GPT is based on left-to-right attention, where every position can only attend to previous positions in the masked multi-head attention layer of each Transformer Block. Transformer-ED first encodes dialogue history, and then the final outputs of encoder are fed into the decoder to generate a response. The encoder uses bi-directional attention, while the decoder uses left-to-right attention.

In this framework, the decoder is stacked upon the encoder outputs. 
This makes the fine-tuning process less effective in updating the encoder's parameters. The potential problem with separate encoder and decoder has been noticed in a previous work on abstractive summarization \cite{liu2018generating}, which noticed that an explicit encoder in the Transformer-ED architecture might be redundant, i.e. the encoding step could be directly incorporated in the decoder, allowing for a more direct update of the parameters. Our experiments will confirm this problem. 

\subsection{Transformer-Dec}
\label{sec:trans_dec}
Transformer-Dec is a decoder-only architecture utilizing GPT. It encodes dialog history using only left-to-right attention, in the same way as in GPT pre-training. However, the fact that only left-to-right attention is used may limit the scope of context when generating responses. Many previous studies have shown that the context after the token is useful for the encoding of the token. It might be better to use bi-directional attention whenever possible.

\subsection{Transformer-MLM and AR}
\label{sec:trans_mlm}
Transformer-MLM and Transformer-AR have an identical decoder-only architecture \footnote{Since the architecture is not explicit encoder-decoder, we categorize it into decoder-only though BERT is usually viewed as a pre-trained encoder.} that employs different type embeddings and self-attention masks for the source/encoder and target/decoder sides: they use bi-directional attention on the source side and left-to-right attention on the target side.

The pre-trained BERT \cite{devlin2018bert} is a bi-directional architecture using MLM as the pre-training objective. The same attention mechanism is used in the encoder part of Transformer-MLM/AR, but the decoder part uses left-to-right attention. 
On the loss function, Transformer-MLM fine-tunes model parameters using MLM as BERT, which masks a certain percentage of tokens at the target side and tries to predict them; while Transformer-AR uses a different auto-regressive objective, which tries to predict the next tokens successively. 

\subsection{Applications of the frameworks}
\label{sec:application}
The frameworks we described have been recently applied to dialogue generation. For personalized response generation, \citet{wolf2019transfertransfo} uses Transformer-Dec and \citet{zheng2019pre} utilizes Transformer-ED. \citet{lin2019caire} uses Transformer-Dec for empathetic response generation. The training data for fine-tuning could be limited in practice. Several studies propose to further pre-train models using large-scale dialogue datasets before fine-tuning: \citet{zhang2019dialogpt} trains Transformer-Dec on 147M Reddit data, \citet{dong2019unified} trains Transformer-MLM on natural language understanding and natural language generation datasets, \citet{shuster2019dialogue} trains Transformer-AR on large-scale Reddit data and then jointly trains on 12 dialog sub-tasks, and \citet{bao2019plato} trains a variant of Transformer-AR on large-scale Reddit and Twitter data. Another trend emerging in the literature \cite{adiwardana2020towards, roller2020recipes, bao2020plato} is to train Transformer variants with billions of parameters on extremely large-scale dialogue dataset, e.g. elaborately pre-processed Reddit. 

The purpose of our study is different: we do not intend to augment training data to push the performance. Instead, we want to examine the intrinsic characteristics of different frameworks for dialogue generation by comparing them on the same datasets. In addition, it may often be the case that the training data for fine-tuning is limited in some domains. So, we also need to investigate the behaviors of the frameworks in such situations. Therefore, in the next section, we will compare the 4 frameworks on 3 public datasets, and examine them with two data scales -- millions and 100K.

\section{Experimental Comparison}
\label{sec:experiments}

\begin{table*}[t]
\centering
\small
\begin{tabular}{lllllllll}
\hline
\hline
\textbf{Model} & \textbf{BLEU-1} & \textbf{BLEU-2} & \textbf{BLEU-3} & \textbf{CIDEr} & \textbf{Dist-1} & \textbf{Dist-2}  & \textbf{avgLen}\\
\hline
SEQ2SEQ-MMI & 10.872 (**) & 4.555 (**) & 2.259 (/) &  \textbf{0.119} (/) & 0.008 (**) & 0.028 (**) & 10.6 \\

% \hline

Trans-ED & 15.319 (**) & 4.877 (**) & 2.037 (**) &  0.097 (**) & 0.014 (**) & 0.063 (**) & 19.0 \\

Trans-Dec & 14.363 (**)  & 4.861 (**)  & 2.120 (*) &  0.101 (**) & \textbf{0.031} (**) & \textbf{0.178} (/) & 19.9 \\

Trans-MLM & 13.749 (**) & 4.253 (**) & 1.715 (**) & 0.061 (**) & 0.018 (**) & 0.106 (**) & 29.3\\

Trans-AR & \textbf{15.694} & \textbf{5.221} & \textbf{2.272} & \textbf{0.119} & 0.029 & 0.164 & 18.9 \\

\hline
\hline

Trans-ED & 14.813 (**) & 4.249 (**) & 1.330 (**) & 0.066(**) & 0.001 (**) & 0.004 (**) & 18.4 \\

Trans-Dec & 13.805 (**) & 4.407 (/) & 1.787 (/) & 0.092(/) & \textbf{0.033} (**) & \textbf{0.195} (**) & 20.2 \\

Trans-MLM & 15.487(/) & 4.766(/) & 1.814(/) & 0.092 (/)  & 0.016(**) & 0.080(**) & 19.7 \\

Trans-AR & 15.213 (**) & 4.700 (/) & 1.767 (/) & 0.090(*) & 0.019(**) & 0.091(**) & 18.8 \\

\hline

PF-free & 15.880 & 4.970 & 1.868 & 0.093 & 0.022 & 0.114 & 15.7 \\
FG-free & \textbf{16.395} & \textbf{5.218} & \textbf{2.043} & \textbf{0.101} & 0.026 & 0.129 & 16.2 \\

PF\&FG-free & 15.714 & 4.916 & 1.780  & 0.093 & 0.020 & 0.111 & 18.4 \\

\hline
\end{tabular}
\caption{\label{tab:twitter} Evaluation results on Twitter dataset. The results of the second part is of the 100k data setting. PF-free denotes the method with reduced pretrain-finetune discrepancy of Transformer-MLM. FG-free denotes the method that removes finetune-generation discrepancy of Transformer-MLM. Two-sided t-test compares each method with the one without () sign, which is usually the best performer.}
\end{table*}

%%%%Table
\begin{table*}[t]
\centering
\small
\begin{tabular}{lllllllll}
\hline
\hline
\textbf{Model} & \textbf{BLEU-1} & \textbf{BLEU-2} & \textbf{BLEU-3} & \textbf{CIDEr} & \textbf{Dist-1} & \textbf{Dist-2} & \textbf{avgLen}\\ 
\hline

SEQ2SEQ-MMI & 12.056(**) & 5.512(**) & 2.841(**) & 0.142(**) & 0.005(**) &  0.024(**)  &  9.8 \\

HRED-MMI & 13.518(**) & 4.564(**) & 1.947(**) & 0.060(**) & 0.001(**) & 0.003(**) & 13.6 \\

Trans-ED & 19.295(/) & 6.712(**) & 2.986(*) &  0.125(**) & 0.010(**) & 0.069(**) & 16.8 \\

Trans-Dec & 18.974(*) & 6.911(/) & 3.022(*) & 0.130(**) & 0.018(**) & 0.134(**) & 18.0 \\

Trans-MLM  & 17.574(**) & 5.884(**) & 2.552(**) & 0.096(**) & 0.012(**) & 0.097(**) & 25.5 \\

Trans-AR & \textbf{20.103} & \textbf{7.270} & \textbf{3.339} & \textbf{0.143} & 0.017 & 0.127 & 16.8  \\

\hline
\hline
Trans-ED & 14.195(**) & 4.533(**) &  1.756(**) & 0.074(**) & 0.003(**) &  0.012(**) & 16.3 \\

Trans-Dec & 17.944(**) & 6.360(/) & 2.727(/) & 0.121(*) & \textbf{0.018}(**) & \textbf{0.143}(**) & 18.3 \\

Trans-MLM & 18.338(**) & 6.018(**) & 2.480(**) & 0.108(**)  & 0.011(**) & 0.066(**) & 17.0 \\

Trans-AR & 19.005 (/) & 6.431 (/) & 2.733 (/) & 0.114(*) & 0.012(/) & 0.078(**) & 17.4 \\

\hline

PF-free & 19.116 & 6.356  & 2.684  & 0.118 & 0.012 &  0.086 &  16.7 \\

FG-free & 18.884 & \textbf{6.530} & \textbf{2.869} & \textbf{0.125} & 0.014 & 0.095 & 17.3 \\

PF\&FG-free & \textbf{19.024} & 6.448 & 2.740 & 0.118 & 0.012 & 0.087 & 17.1 \\

\hline 
\end{tabular}
\caption{\label{tab:ubuntu} Evaluation results on Ubuntu dataset. The results of the second part is of the 100K data setting.}
\end{table*}

%%%% Table 
\begin{table*}[t]
\centering
\small
\begin{tabular}{lllllllll}
\hline
\hline
\textbf{Model} & \textbf{BLEU-1} & \textbf{BLEU-2} & \textbf{BLEU-3} & \textbf{CIDEr} & \textbf{Dist-1} & \textbf{Dist-2} & \textbf{avgLen}\\ 
\hline

SEQ2SEQ-MMI & 15.550(**) & 6.814(**) & 3.321(**) & 0.168(**) & 0.011(**) & 0.036(**) & 11.2 \\

HRED-MMI & 13.278(**) & 3.845(**) & 1.398(**) & 0.047(**) & 0.001(**) & 0.003(**)  & 13.8 \\

Trans-ED & 17.946(/) & 6.626(**) & 3.213(**) & 0.165(**) & 0.039(**) & 0.203(**) & 18.8 \\

Trans-Dec & 17.581(**) & 6.790(*) & 3.372(*) & 0.180(**) & 0.043(/) & \textbf{0.248}(**) & 18.2 \\

Trans-MLM & 18.672(**) & 7.115(**) & 3.484(/) & 0.177(**) & 0.041(**) & 0.215(**) & 16.8 \\

Trans-AR & 18.849 & 7.245 & \textbf{3.662} & \textbf{0.192} & \textbf{0.044} & 0.235 & 16.8 \\

\hline
\hline
Trans-ED & 17.337(**) & 5.366(**) & 1.967(**) & 0.073(**) & 0.001(**) & 0.003(**) & 17.1 \\

Trans-Dec & 17.460(**) & 6.586(**) & 3.161(**) & 0.172(**) & \textbf{0.045}(**) & \textbf{0.254}(**) & 17.7 \\

Trans-MLM & 19.193 (/) & 6.877 (/) & 3.175(/) & 0.152(**) & 0.029(**) & 0.128(**) & 15.0 \\

Trans-AR & 18.749(/) & 6.746(/) &  3.119(*) & 0.153(**) & 0.031(**) & 0.141(**) & 16.2 \\

\hline

PF-free & 18.466 & 6.688 & 3.075 & 0.169 & 0.038 & 0.180 & 14.1 \\

FG-free & 18.610 & \textbf{6.937} & \textbf{3.302} & \textbf{0.175} & 0.040 & 0.191 & 14.1 \\

PF\&FG-free & \textbf{19.302} & 6.923 & 3.073 & 0.159 & 0.034 & 0.164 & 15.3 \\

\hline
\end{tabular}
\caption{\label{tab:reddit} Evaluation results on Reddit dataset. The results of the second part is of the 100k data setting.}
\end{table*}

\subsection{Datasets}
We use three public datasets that are widely used for open-domain dialog generation. Some important characteristics of the datasets are summarized in Table \ref{tab:dataset}. For each dataset, we also evaluate model performance using only 100K training data with the same validation set and test set, in order to simulate the situation of limited data. Many dialogue datasets that have human annotations such as PersonaChat \cite{zhang2018personalizing} and EmpatheticDial \cite{rashkin2019towards} are in similar small scale. 

\begin{table}[h]
\small
\centering
\begin{tabular}{l|lll}
\hline 
\hline
 & Twitter & Ubuntu & Reddit \\
\hline 
Dialog Turns & 2 & 3 & 3 \\
Avg. Src Len & 18 & 34 & 28  \\
Avg. Tgt Len & 16 & 16 & 15  \\
\hline 
Train Set & 2M & 1.5M & 3M \\
Valid Set & 60K & 30K & 80K \\
Test Set & 20K & 20K & 20K \\
\hline
\end{tabular}
\caption{\label{tab:dataset} Key characteristics of the three public datasets. For each dataset, we also evaluate model performance using 100K training data and the same test set.}
\end{table}

\textbf{Twitter Dialog Corpus}  \footnote{https://github.com/Marsan-Ma-zz/chat\_corpus} is collected from Twitter consisting of 2.6M (message, response) pairs. We filtered out a sample if its history length is longer than 72 words or shorter than 6 words. Samples whose response is longer than 36 words or shorter than 6 words are also removed. As results, we keep 2M samples. 

\textbf{Reddit Conversational Corpus} \footnote{https://github.com/nouhadziri/THRED}\cite{dziri2019augmenting} is a high-quality 3-turns conversational dataset collected from 95 selected subreddits. The dataset contains 9.2M samples. We applied the same length constraints and then randomly sampled 3M samples.  

\textbf{Ubuntu Dialogue Corpus V2.0}  \footnote{https://github.com/rkadlec/ubuntu-ranking-dataset\\-creator} \cite{lowe2017training} is two-person conversations extracted from the Ubuntu chat logs of technical support for various Ubuntu-related problems. We split a conversation that has more than 3 turns into several 3-turn dialogues. 
Besides, we filtered out samples to have similar  dialogue history and response length as the other two datasets.

\subsection{Implementation Details}
We equip all transformer variants with an identical decoding algorithm\footnote{From UniLM, https://github.com/microsoft/unilm/} to avoid extra factor affecting the generation quality, which uses beam search with beam size of 4, prevents duplicated uni-grams, and sets minimum response length that encourages diverse generation \cite{roller2020recipes}. 
The minimum response length is set to make the average length of generated responses match with the average target length of the dataset. 
Generation results are evaluated after applying an identical word tokenization method. With two P100 GPU devices, we fine-tune all models for 4 epochs on Twitter and Reddit dataset, and for 5 epochs on Ubuntu dataset. Maximum input length is set to 128. Our methods (PF-free and FG-free, which will be described in Section \ref{sec:my_methods}) do not add parameters or increase runtime in comparison to Transformer-MLM. More implementation details are given in Appendix \ref{app:imp}. 

\subsection{Evaluation}
\paragraph{Metrics} We compare the similarity between generated responses and ground-truth responses using\footnote{We use an open-source evaluation tool: https://github.com/Maluuba/nlg-eval}: \textbf{BLEU} \cite{papineni2002bleu} evaluating how many n-grams (n=1,2,3) overlapped; \textbf{CIDEr} \cite{vedantam2015cider} utilizing TF-IDF weighting for each n-gram. Besides, we evaluate response diversity using \textbf{Distinct} (denoted Dist) \cite{li2016diversity} that indicates the proportion of unique n-grams (n=1,2) in the entire set of generated responses. Notice that only when similarity scores are comparable that higher Distinct scores might represent better responses. Otherwise, grammatically wrong responses that are not fluent could also have high Distinct scores. We also report average length of the generated responses (denoted \textbf{avgLen}). We run statistical significance tests using two-sided t-tests. 
Scores are denoted with * ($p < 0.05$) or ** ($p < 0.01$) for statistically significant differences. 

\paragraph{Human Evaluation} Furthermore, we ask human evaluators to rate a response in $\{0, 1, 2\}$. 2 represents a coherent and informative response. Details are given in Appendix \ref{app:human}. We also do a pair-wise evaluation to compare two models and indicate which one is better. To reduce time cost, we only evaluate model performances for the Twitter and Reddit datasets that are closer to daily dialogue. However, during evaluation, we observe that $\sim 65\%$ Reddit data are professional discussions that are difficult to understand given only 2-turn dialog history. The percentage is $\sim 30\%$ for Twitter data. We skip these test samples, and at the end the test set for each dataset consists of 200 random samples. The inter-rater annotation agreement is measured using the Cohen’s kappa \cite{cohen1960coefficient}. The $\kappa$ is $0.44$ and $0.42$ on average for Twitter and Reddit, which indicates moderate agreement.

\begin{table}[h]
\centering
\small
\begin{tabular}{lll}
\hline
\hline
\textbf{Model} & \textbf{Score} (M) & \textbf{Score} (K) \\
\hline
SEQ2SEQ-MMI & 0.39 (**) & - \\
Trans-ED & 0.53 (**) & 0.11 (**) \\
Trans-Dec & 1.26 (/) & 1.03 (/) \\
Trans-MLM & \textbf{1.28}  & 1.03 (/) \\
Trans-AR & 1.20 (/) & 0.84 (*) \\
PF\&FG-free & 1.22 (/) & \textbf{1.16} \\
\hline
\hline
& Trans-MLM (M) & PF\&FG-free  (K) \\
\hline
SEQ2SEQ-MMI & (7\%, 63\%) & - \\
Trans-ED & (10\%, 60\%) & (3\%, 65\%)\\
Trans-Dec & (26\%, 37\%) & (28\%, 41\%)\\
Trans-MLM & / & (27\%, 37\%) \\
Trans-AR & (27\%, 40\%) & (24\%, 43\%)\\
PF\&FG-free  & (26\%, 34\%) & / \\
\hline

\end{tabular}
\caption{\label{tab:humanscore} Human evaluation including pair-wise evaluation for generated response quality for million-scale (M) Twitter dataset and its 100K training subset (K). }
\end{table}

\begin{table}[h]
\centering
\small
\begin{tabular}{lll}
\hline
\hline
\textbf{Model} & \textbf{Score} (M) & \textbf{Score} (K) \\
\hline
SEQ2SEQ-MMI & 0.12 (**) & - \\
Trans-ED & 0.33 (**) & 0.03 (**) \\
Trans-Dec & 0.72 (/) & \textbf{0.71} (/) \\
Trans-MLM & 0.72 (/)  & 0.32 (**) \\
Trans-AR & \textbf{0.77} & 0.28 (**)\\
PF\&FG-free & 0.76 (/) & 0.59 \\
\hline
\hline
 & Trans-AR (M) & PF\&FG-free (K) \\
\hline
SEQ2SEQ-MMI & (6\%, 46\%) & - \\
Trans-ED & (8\%, 38\%) & (0\%, 41\%)\\
Trans-Dec & (17\%, 21\%) & (32\%, 27\%)\\
Trans-MLM & (25\%, 26\%) & (13\%, 34\%) \\
Trans-AR & / & (12\%, 32\%)\\
PF\&FG-free & (26\%, 27\%) & /  \\
\hline

\end{tabular}
\caption{\label{tab:humanscore2} Human evaluation including pair-wise evaluation for generated response quality for million-scale (M) Reddit dataset and its 100K training subset (K).}
\end{table}

\subsection{Architecture Analysis} \label{sec:framew_comp}

\textbf{Transformer-ED} performs the worst. We observe that it is the only framework that tends to generate safe responses such as "i am not sure what you are talking about" (except on million-scale Reddit dataset). 
This phenomenon is also revealed by its substantially lower Distinct scores. 
When training with 100K data, the generated responses are almost bland, which can be observed in generation samples (Appendix \ref{app:cases}). As mentioned earlier (\cref{sec:trans_ed}), the Transformer-ED architecture might be redundant \cite{liu2018generating} -- the separate encoder may not be necessary and be difficult to optimize. 

\textbf{Transformer-AR} obtains the highest BLEU and CIDEr scores on all three million-scale datasets. Since the only difference from Transformer-Dec is that Transformer-AR applies bi-directional attention on source side, we can explain the difference in performance by the attention mechanism. This suggests that bi-directional attention on the source side helps the model to better encode the dialogue history and thus produce better responses.
Human evaluation results also show that Transformer-AR performs well when fine-tuning with million-scale data. 
However, with small-scale datasets, Transformer-AR loses its advantages, and becomes less effective than Transformer-Dec and Transformer-MLM. We believe that this is due to the discrepancies in Transformer-AR, which will be analyzed later.

\textbf{Transformer-Dec} is able to generate the most diverse responses. We attribute this to the left-to-right attention on the source side, which can introduce more flexibility for generation since it does not have constraints from the right side. 
Furthermore, when fine-tuning data are limited, Transformer-Dec has the best performance on both Twitter and Reddit datasets according to human evaluation.

\textbf{Transformer-MLM} outperforms other frameworks in terms of BLEU metrics when fine-tuning data are limited, which might have benefited from the bi-directional attention on the source side. 
Although with million-scale dataset its automatic evaluation scores are not the best, human evaluation shows that its performance is comparable to Transformer-Dec and Transformer-AR. 
Furthermore, we observe that Transformer-Dec and sometimes Transformer-AR will simply completely copy the input sentence in the output (see  examples in Appendix \ref{app:cases}, Table \ref{tab:twittercase3} and \ref{tab:redditcase1}), but this seldom happens with Transformer-MLM. We conjecture that the MLM objective contributed in alleviating the issue.

Overall, our experiments suggest model architecture is strongly related to the success of fine-tuning a pre-trained model for dialogue generation.
The best configuration uses bi-directional attention on the source side and combines encoder and decoder in the same multi-layer transformer blocks.

\subsection{Discrepancy Impact}
In Table \ref{tab:frameworks}, we emphasized the pretrain-finetune discrepancy in red. All the frameworks, except Transformer-Dec, have some discrepancies, while only Transformer-MLM has finetune-generation discrepancy: during training, the model input has random masks as shown in Figure \ref{Fig:model}, while in generation process, the input consisting does not contain masks. Discrepancies may affect the model performance since models with such discrepancies cannot best exploit the pre-trained language model or best employ the fine-tuned model. In this subsection, we try to understand how model discrepancies affect the model.

When a large training dataset is available for fine-tuning, we observe the impact of pretrain-finetune discrepancy is less severe -- the model can gradually be adapted to the given task. However, if the training data are limited, the discrepancy problems may surface. 

Specifically, Transformer-ED is the framework that has the largest pretrain-finetune discrepancy. We observe that its performance decreases a lot from large training data to small training data. On the other hand, Transformer-Dec has the least pretrain-finetune discrepancy, and we can observe much smaller decrease in performance (especially in human evaluation). We can also compare Transformer-MLM and Transforme-AR, which use an identical architecture. Transformer-AR has additional pretrain-finetune discrepancy due to its auto-regressive objective in fine-tuning. We can see that the performance of Transformer-AR is more reduced when fine-tuned on a small dataset: Transformer-AR performs better than Transformer-MLM with large training datasets, but loses its advantages with small training datasets according to automatic evaluation. The human evaluation can show this larger decrease more clearly.

The above analysis suggests that with a small dataset one should make efforts to reduce pretrain-finetune discrepancy to best exploit pre-trained models. In terms of finetune-generation discrepancy, since this is a problem for Transformer-MLM only, it is difficult to show its impact because no comparison with other frameworks is possible. 

As we have revealed the two discrepancies as potential factors that negatively affect the model performance, in the next section, we propose approaches to reduce pretrain-finetune discrepancy and finetune-generation discrepancy, and we will test the performance of the modified models.

\section{Discrepancy-Free Transformer-MLM}
As we have shown that the architecture of Transformer-MLM and Transformed-AR is the best in Section \ref{sec:framew_comp}, our question now is how the model performance would be if we reduce its discrepancies.

\subsection{Multi-Layer Transformer}
\label{sec:maskattn}
Let us first briefly introduce the general transformer mechanism, and then describe our methods to reduce discrepancies.

\begin{figure}[t]
\centering
\includegraphics[height=1.6in]{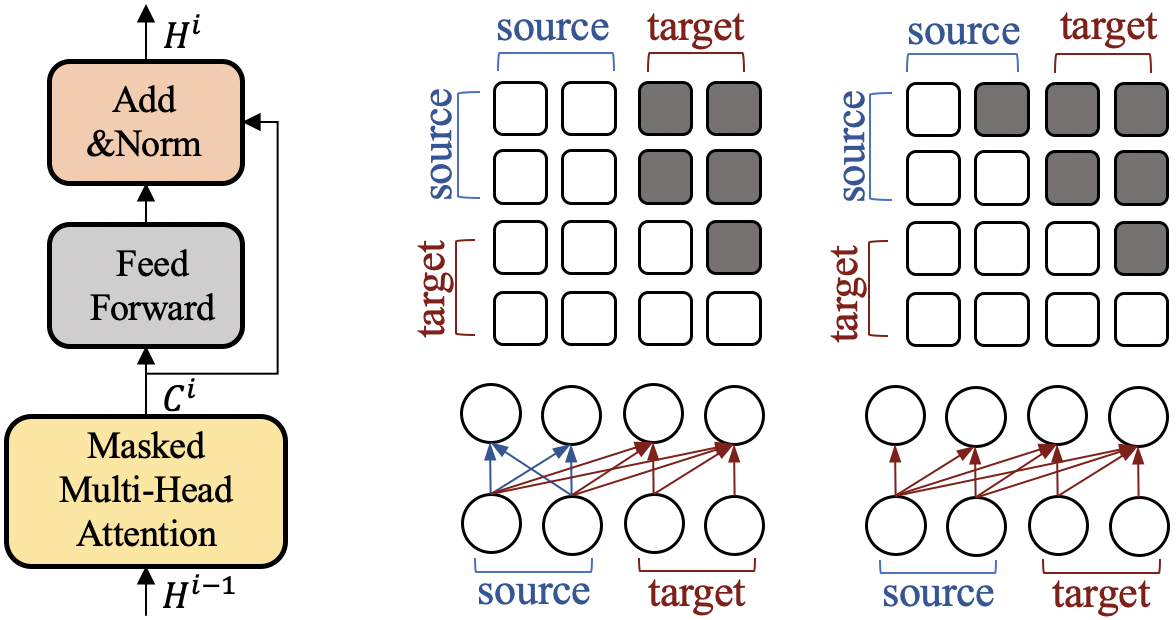}
\caption{$i$-th Transformer Block and two $\mathbf{M}$ settings represented in two ways. Shaded areas are blocked.}
\label{Fig:mask}
\end{figure}

A dialog history is denoted by $x$, and a corresponding response is denoted by $y$. The input to the multi-layer transformer is the concatenation of dialog history and the response. When using MLM objective, the response is randomly masked. The input representation $\mathbf{H}^0 \in \mathbb{R}^{n \times d_h}$, where $n$ is the input length and $d_h=768$ is the hidden dimension, is the sum of token embedding, position embedding, and type embedding at each position. The type embeddings introduce a separation between encoder/source side and decoder/target side in order to warrant different treatments in the model. Then, $\mathbf{H}^0$ is encoded into hidden representations of $i$-th layer $\mathbf{H}^i=[\mathbf{h}_1^i, ...,\mathbf{h}_{n}^i]$ by: $\mathbf{H}^i={\rm Trans}^i(\mathbf{H}^{i-1}), \quad i \in [1,L]$, where ${\rm Trans}^i$ denotes the $i$-th Transformer Block as shown in Figure \ref{Fig:mask}(Left). The core component of a transformer block is the masked multi-head attention, whose outputs are $\mathbf{C}^i=[\mathbf{c}_1^i, ...,\mathbf{c}_{n}^i]$ that are computed via $\mathbf{C}^i = {\rm Concat}(\mathbf{head}_1, ..., \mathbf{head}_h)$, with
\begin{equation}
\mathbf{head}_j = {\rm softmax}(\frac{\mathbf{Q}_j\mathbf{K}_j^{T}}{\sqrt{d_k}}+\mathbf{M})\mathbf{V}_j
\end{equation}
where $\mathbf{Q}_j, \mathbf{K}_j, \mathbf{V}_j \in \mathbb{R}^{n\times d_k}$ are obtained by transforming $\mathbf{H}^{i-1} \in \mathbb{R}^{n\times d_h}$ using $\mathbf{W}_{j}^{Q}, \mathbf{W}_j^{K}, \mathbf{W}_j^{V} \in \mathbb{R}^{d_h\times d_k}$ respectively. $\mathbf{M} \in \mathbb{R}^{n\times n}$ is the self-attention mask matrix that determines whether a position can attend to other positions. $\mathbf{M}_{ij} \in \{0, -\infty \}$. In particular, $\mathbf{M}_{ij}=0$ allows the $i$-th position to attend to $j$-th position and $\mathbf{M}_{ij}=-\infty$ prevents from it. Figure \ref{Fig:mask} (Right) shows two $\mathbf{M}$ settings that are applied by Transformer-MLM/AR and Transformer-Dec respectively.

\subsection{Pretrain-Finetune Discrepancy} 
\label{sec:my_methods}

If we reduce the pretrain-finetune discrepancies of Transformer-AR, we get Transformer-MLM by replacing the AR objective to MLM objective. The discrepancy of Transformer-MLM comes from the left-to-right attention on the target side that has not been pre-trained in BERT. Therefore, this discrepancy cannot be eliminated during fine-tuning for a generation task. However, we might alleviate the discrepancy by using bi-directional attention also on the target side. Specifically, at inference time, to generate a new token denoted as $g_t$, [MASK] is fed into $t$-th position, denoted as $g_t$-M. Previously generated tokens $g_{<t}$ could be viewed as a special type of dialogue history, and we can apply bi-directional attention on it.  

However, previously left-to-right attention is used on $g_{\leq t}$, and thus only hidden states at $(t-1)$-th and $t$-th positions need to be updated. If we directly apply bi-directional attention on $g_{\leq t}$, all hidden states at $\leq t$ positions need to be updated. In other words, this generation strategy requires to update all hidden states at each time step of generation. 
Our experiments show that with such high updating frequency the generated responses usually lack fluency (e.g. BLEU-2 is 1.519 on Twitter). Furthermore, to keep training consistent with inference, only one token can be masked in each training sample; otherwise, there will be conflict for the self-attention mask (Appendix \ref{app:conflict}, Figure \ref{Fig:biattn_train}), i.e. different self-attention mask matrices are required for different masked tokens, while only one mask matrix can be provided per training sample. This would lead to much lower training efficiency: the loss on validation set only decreases slightly to $5.39$ from $6.27$ after four epochs, while 
Transformer-MLM masking 40\% of the target tokens can reduce it to $4.35$. 

To summarize, in generation, we cannot always update previous hidden states using bi-directional attention, and the training process should be consistent to the generation in order not to add new finetune-generation discrepancy. Therefore, we propose a simple method by setting a time-step interval for bi-directional attention on the target side to decrease updating frequency -- within the interval we apply left-to-right attention and at the end of an interval we apply bi-directional attention. The corresponding training method allows us to mask multiple target tokens at the same time to guarantee training efficiency.

\begin{figure}[t]
\centering
\includegraphics[height=1.2in]{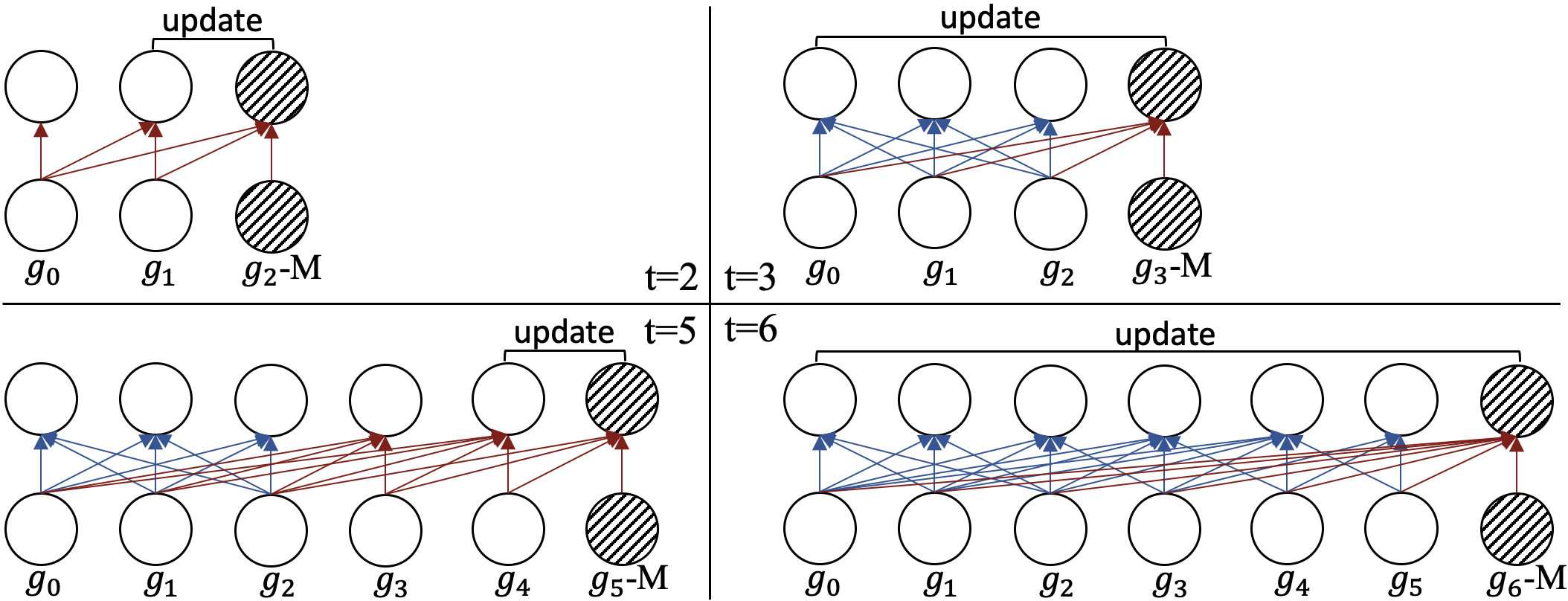}
\caption{Alleviate pretrain-finetune discrepancy of Transformer-MLM by using bi-directional attention on the target side. We show the generation process at 4 different steps and annotate the range of positions to update. Bi-attention interval is 3 in the graph. }
\label{Fig:multis}
\end{figure}

Figure \ref{Fig:multis} illustrates the generation process of our method with interval of 3. Before time step 3, left-to-right attention is used (e.g. t=2). At time step 3, bidirectional attention is allowed. Then left-to-right attention is used (e.g. t=5) before the end of next interval cycle (t=6). Accordingly, the training process is: given a target response, we first randomly select among all (3 in the figure) possible attention patterns (e.g. the case of t=3 or t=5 in Figure \ref{Fig:multis}, where we apply bi-directional attention on $y_{0, 1, 2}$); then in the part of left-to-right attention, we randomly mask several tokens. We can mask multiple tokens because this part applies left-to-right attention and the masks at other positions will not influence the prediction on a given mask. We call our modified approach PF-free, which means that the pretrain-finetune discrepancy is reduced.

\subsection{Finetune-Generation Discrepancy}
\label{sec:obj}

\begin{figure}[t]
\centering
\includegraphics[height=1.4in]{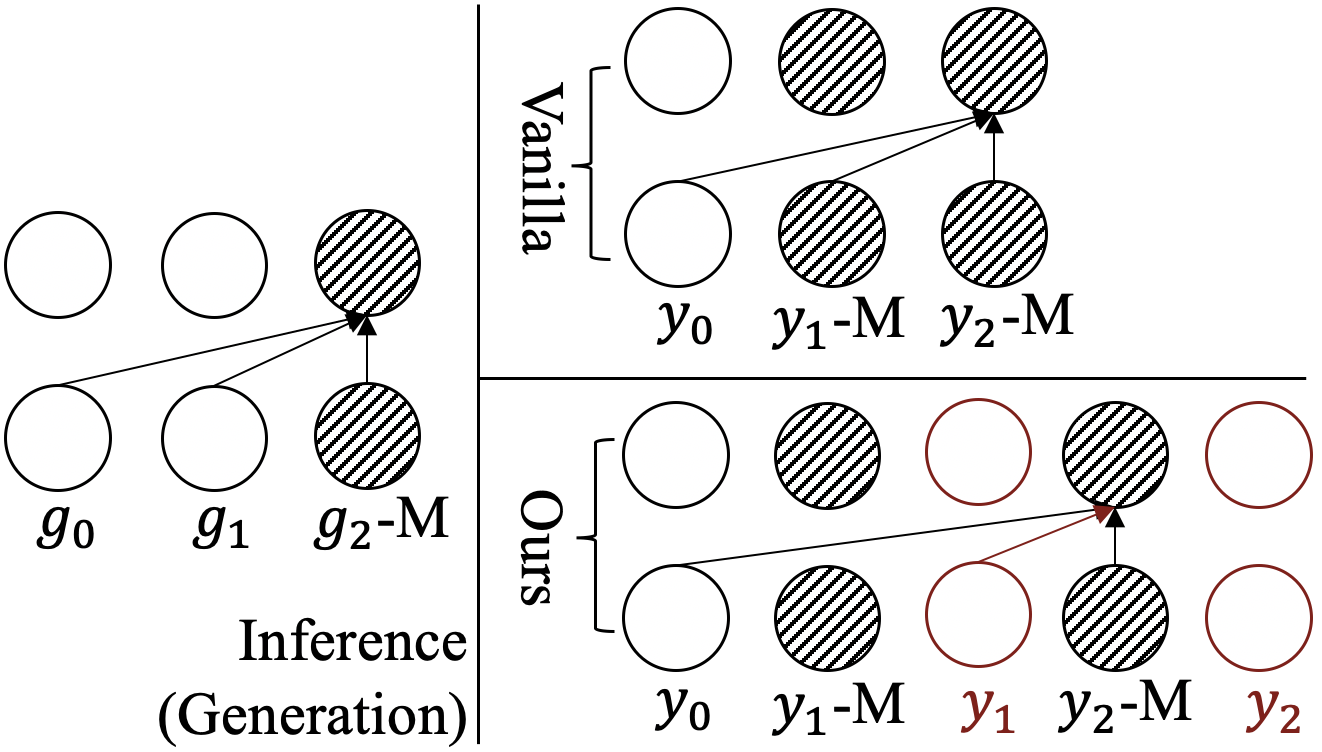}
\caption{Eliminating finetune-generation discrepancy of Transformer-MLM. We only plot the attention connection at $y_2-M$.}
\label{Fig:finegen}
\end{figure} 

A model having finetune-generation discrepancy means the way that it is used in generation (inference/test) is different from the way it has been trained. Only Transformer-MLM has finetune-generation discrepancy because of its MLM objective as shown in Figure \ref{Fig:finegen}: during training, there is a masked token, $y_1$-M, before $y_2$-M, while in generation process, there is not a masked token before when generating the token $g_2$-M. 

Therefore, we propose that at training time, rather than replacing the tokens with [MASK] as in vanilla MLM, we keep all original input tokens unchanged and prepend [MASK] tokens in the input sequence as illustrated. The prepended [MASK] token uses the same position embedding of the corresponding token. Then, every position after $y_1$-M attends to $y_1$ instead of the [MASK] token, and thus the finetune-generation discrepancy of MLM is eliminated. A similar method has also been explored in \citet{bao2020unilmv2}, where they introduced an extra pseudo mask in additional to [MASK] and prepend it before the original token in order to handle factorization steps of their partially auto-regressive language model.  

We call the modified model FG-free, in which the finetune-generation discrepancy is reduced.

\subsection{Experimental Results}

The results with PF-free, FG-free and PF\&FG-free models are reported in previous tables together with other models. We can see that both methods to reduce discrepancies improve Transformer-MLM performance. Specifically, PF-free improves BLEU on Twitter and Ubuntu datasets, and it always outperforms Transformer-MLM in terms of CIDEr and Distinct scores on all three datasets. Besides, as have been discussed, Transformer-MLM that has smaller pretrain-finetune discrepancies outperforms Transformer-AR when training data are small. These results together suggest that the effort to reduce pretrain-finetune discrepancy when training data are small is beneficial.

In comparison, PG-free results in even better performances, which always brings statistically significant improvement over Transformer-MLM in all automatic metrics on all three datasets. Therefore, we can see that it is necessary to apply our method on Transformer-MLM to eliminate finetune-generation discrepancy. However, when we combine the two correction methods to reduce both discrepancies, we do not observe an cumulative effect. The relation between the two discrepancies remains unclear. This question is worth being investigated further in the future.

\section*{Conclusion}
In this paper, we compared experimentally 4 main frameworks using pre-trained language models for dialogue generation. The comparison revealed the best strategy to use pre-trained model: using bi-directional attention, and integrate encoder and decoder in the same transformer blocks. 

In addition to the architectural appropriateness, we also examined the discrepancies of each framework. We identified two discrepancies: pretrain-finetune discrepancy between the pre-training and fine-tuning processes, and finetune-generation discrepancy that exists between the fine-tuning and generation processes.  Our experiments showed that discrepancies affect the model performance since models with such discrepancies cannot best exploit the pre-trained language model or best employ the fine-tuned model. To reduce the negative impact of discrepancies, we proposed two methods to reduce the pretrain-finetune discrepancy and finetune-generation discrepancy respectively, and both improved the model performance. This confirms the necessity to design approaches to dialogue generation based on pre-trained models, containing as little discrepancy as possible.

The appropriate utilization of pre-trained models is important, and is worth being paid more attention to. This study is a first investigation on model discrepancy and more studies are needed. 

\bibliography{anthology,acl2020}
\bibliographystyle{acl_natbib}

\appendix

\section{Attention Conflict Illustration}
\label{app:conflict}

\begin{figure}[h]
\centering
\includegraphics[height=0.9in]{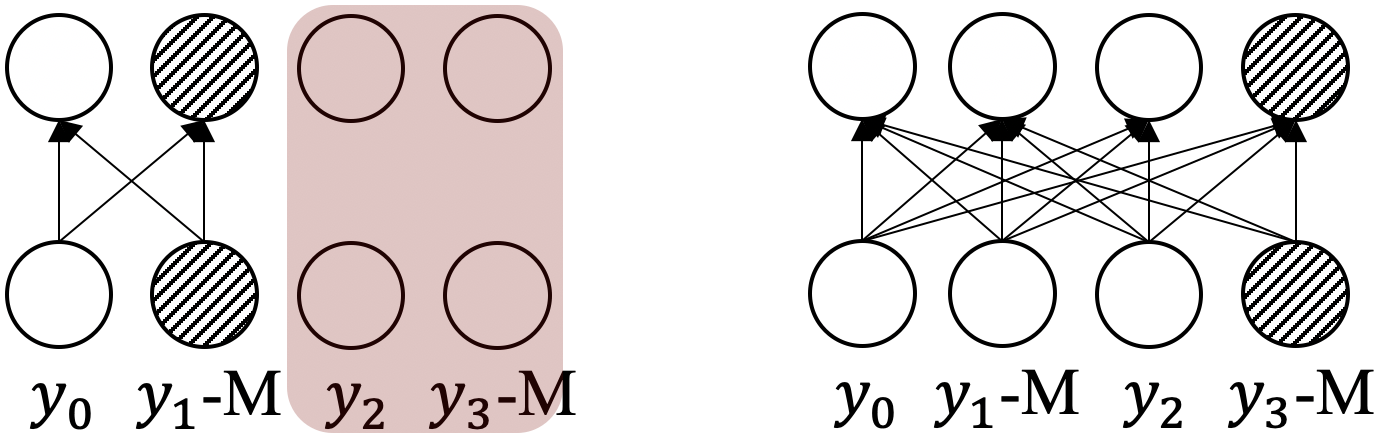}\caption{Self-attention mask, $\mathbf{M}$, conflicts -- if predicting $y_1$, $\mathbf{M}$ is as the left figure, where $y_2$ and $y_3$-M are "future" and forbidden to access by $y_1$-M; if predicting $y_3$, $\mathbf{M}$ is as the right figure, in which case $y_1$-M accesses to $y_2$ and $y_3$-M. Masking two positions thus causes conflicts.}
\label{Fig:biattn_train}
\end{figure}

In generation, it seems appropriate if applying bi-directional attention at each generation step (however, we experimentally show the efficiency and effectiveness issues.). The corresponding training method is extremely inefficient -- only one token at the target side could be masked for each training sample; otherwise there will be attention conflicts. In Figure \ref{Fig:biattn_train}, we assume $y_1$ and $y_3$ are masked and need to be predicted at the same time. 

\section{Baselines}
\label{app:baseline}
We include two general RNN-based frameworks in this comparison to show how pre-trained models perform against them --  \textbf{SEQ2SEQ-MMI} \cite{li2016diversity}, a seq2seq model using bi-directional GRU encoder and applying Maximum Mutual Information (MMI) as the objective function to generate more diverse responses, and \textbf{HRED-MMI} \footnote{https://github.com/hsgodhia/hred}, a hierarchical recurrent encoder-decoder neural network \cite{serban2016building} applying diverse decoding strategy based on MMI \cite{li2016mutual}.

\section{Implementation Details}
\label{app:imp}
We implemented SEQ2SEQ-MMI based on OpenNMT \footnote{http://opennmt.net/}, HRED-MMI based on an open-source implementation\footnote{https://github.com/hsgodhia/hred}, Transformer-ED based on the codes of \citet{zheng2019pre}, Transformer-Dec upon \citet{wolf2019transfertransfo}\footnote{https://github.com/huggingface/pytorch-openai-transformer-lm}, Transformer-MLM and Transformer-AR upon \citet{dong2019unified} \footnote{https://github.com/microsoft/unilm/tree/master/unilm-v1}. Hyper-parameters are set following the original papers. 
We set the bi-directional attention interval of PF-free to $5$. Since the average length of ground-truth responses in the datasets is $\sim 15$, This setting is generally appropriate. 

In Table \ref{tab:runtime}, the average runtime is tested using a 1080Ti GPU device, and the batch size is set to take all of the GPU memories. Notice that the runtime will be influenced by code implementation in addition to model structure. 

\begin{table}[h]
\centering
\begin{tabular}{lll}
\hline
\hline
Model & Params & Runtime(min/M) \\
\hline 
SEQ2SEQ-MMI & 66M & 50 \\
HRED-MMI & 58M & 25 \\
Transformer-ED & 117M & 180 \\
Transformer-Dec & 124M & 290 \\
Transformer-MLM & 110M & 140 \\
Transformer-AR & 110M & 140 \\
PF\&FG-free & 110M & 140 \\
\hline
\end{tabular}
\caption{\label{tab:runtime} The number of parameters of each tested approach and the average runtime (minutes) for every million training samples.}
\end{table}

\section{Human Evaluation Details}
\label{app:human}
Some existing studies considered response fluency, coherence, and informativeness. We make the manual evaluation simpler and
ask the human evaluators to rate a response in $\{0, 1, 2\}$. A score of 0 represents an unacceptable response, which might have flaw in fluency and logic or be incoherent. A special case is a bland response such as ``i do not know what you are talking about , but it 's a good point .''. 
A score of 1 represents an acceptable response that might be simple or generic. 2 represents a coherent and informative response. Each generated response is rated by three annotators. Annotators are unaware of which model generates a response. 

\section{Generation Samples}
\label{app:cases}

\begin{table*}[h]
\small
\begin{tabular}{ll}
\hline
\hline
\textbf{Dialog History} & one week before election day , early voting is nearly twice as high as $<$num$>$  \\
\textbf{Gold Response} & i hope it 's $<$person$>$ out in full force . \\
\hline
SEQ2SEQ-MMI & i am not voting for a third party candidate . (2)\\

Trans-ED & i am not sure what you are talking about , but it 's a good thing that $<$person$>$ is going to win . (2) \\

Trans-Dec & that 's not true . early voting in nyc is times higher than the national average (2) \\

Trans-MLM  & it 's not even close to $<$num$>$ \% of the vote . i am sure you are right , but there is a huge difference  \\
& between early voting and $<$person$>$ voter suppression in ca (2) \\

Trans-AR  &  it 's not that high , but i am sure there will be a lot of voter fraud . (2) \\

PF\&FG-free & it 's not even sure how many people voted for $<$person$>$ . $<$num$>$ \% of the vote went to $<$person$>$ (2) \\

\hline
Trans-ED (K) & i am not sure what you are talking about . it 's the only one who is going to do with this (0) \\

Trans-Dec (K) & it 's not that early voting is a bad thing , but it shouldnot be the case . (0) \\

Trans-MLM (K) & it 's a little too late . i am not sure if that is the case , but we will see what happens next (2) \\

Trans-AR (K) & it 's not even close to $<$num$>$ \% of voter turnout . i am sure you are right about that (0)\\

PF\&FG-free (K) &  it 's not even $<$num$>$ \% of the vote . i  am going to be a voter (2) \\

\hline
\end{tabular}
\caption{\label{tab:twittercase1} Responses generated by baselines and our methods trained on the \textbf{Twitter} dataset(million-scale and 100K). Human evaluation scores are given at the end of each generated reply.}
\end{table*}

\begin{table*}[h]
\small
\begin{tabular}{ll}
\hline
\hline
\textbf{Dialog History} & i think about this man every day \\
\textbf{Gold Response} & it is so hypnotic . this man was found dead post-election \\
\hline
SEQ2SEQ-MMI & i do not know what you are talking about . (0)\\

Trans-ED & i think he 's the only one who has a lot of work to do with it . (1) \\

Trans-Dec & i think about this man every day too . he 's my favorite person in the world (2) \\

Trans-MLM  & he 's such a good guy . i love him so much, but it doesnot make sense to me that this is the only way \\
& we can be together in our lives (2) \\

Trans-AR  &  i think about this man every day too . he 's so good at it ! (2) \\

PF\&FG-free & he 's the best friend of mine and i am so  happy for him . it makes me want to cry (2) \\

\hline
Trans-ED (K) & i am not sure what you are talking about , but it 's a good idea . (0)\\

Trans-Dec (K) & i think about this man every day . he 's my hero and i love him (2) \\

Trans-MLM (K) & i think he 's a great actor , but no one cares about him . lol (2) \\

Trans-AR (K) & i think he 's a good guy , but not sure what you are talking about . (0) \\

PF\&FG-free (K) &  i think he is little silly , but you are a fool  . (2) \\

\hline
\end{tabular}
\caption{\label{tab:twittercase2} Responses generated by baselines and our methods trained on the \textbf{Twitter} dataset(million-scale and 100K). Human evaluation scores are given at the end of each generated reply.}
\end{table*}

\begin{table*}[h]
\small
\begin{tabular}{ll}
\hline
\hline
\textbf{Dialog History} & so i asked the person who rang me up if y'all had a comment box . she pointed out the \\
& number on the receipt \& nicely asked me what was up . \\
\textbf{Gold Response} & i asked her why you folks sell those costumes since they are seen as really offensive . \\
& i thought for sure she 'd say , " sorry , i know " \\
\hline
SEQ2SEQ-MMI & i do not know what you are talking about . (0)\\

Trans-ED & i am so sorry for your loss . you are the one who did not have a problem with me (0) \\

Trans-Dec & \textcolor{red}{so i asked the person who rang me up if y'all had a comment box . she pointed out that } \\
& \textcolor{red}{it was not on my receipt . (0)} \\

Trans-MLM  & and i asked her if y ' all had a comment box . she told me that \\
&  it was not my phone number , but the receipt said they were going to call you at $<$num$>$ (0) \\

Trans-AR  & \textcolor{red}{so i asked the person who rang me up if y'all had a comment box . (0)} \\

PF\&FG-free & i was like , " what 's wrong with a comment box . and she did not answer the phone \\
& until $<$num$>$ minutes after we got home from work (2) \\

\hline
Trans-ED (K) & i am not sure what you are talking about , but it 's a good idea . (0)\\

Trans-Dec (K) & i did not have a comment box , so i asked the person who rang me up if y'all had . (0) \\

Trans-MLM (K) & that 's why i do not even know what you are talking about . lol ! (0) \\

Trans-AR (K) & i think it 's a good idea to call you up , but that would be great . (0) \\

PF\&FG-free (K) &  i am not sure if she had a comment box , but it 's probably the same . (2) \\

\hline
\end{tabular}
\caption{\label{tab:twittercase3} Responses generated by baselines and our methods trained on the \textbf{Twitter} dataset(million-scale and 100K). This example shows that Transformer-Dec and Transformer-AR sometimes completely copy the input as the output. }
\end{table*}

%%%%%%%%%% Reddit

\begin{table*}[h]
\small
\begin{tabular}{ll}
\hline
\hline
\textbf{Dialog History} & how old are you guys ? you did not have mechanical pencils ? [SEP] <url> young are you ? \\
& i was in school all the way up to a few years ago and most of us still used non - mechanical ones . \\
\textbf{Gold Response} & i started using mechanical pencil in elementary school about 15 years ago . \\
\hline
SEQ2SEQ-MMI & i do not know what you are talking about . (0)\\

Trans-ED & how old are you ? i was in school all the way up to a few years ago . (1) \\

Trans-Dec & \textcolor{red}{i was in school all the way up to a few years ago and most of us still used non - mechanical ones . (0)} \\

Trans-MLM  & i was in school for a few years , and most of us still use mechanical pencils . it 's not the same age  \\
& as you (2) \\

Trans-AR  &  i was in high school when we used to use mechanical pencils . (2) \\

PF\&FG-free & i was in high school and my teacher told me how to use mechanical pencils . (2) \\

\hline
Trans-ED (K) & i do not know what you are talking about , but it 's just a joke . (0)\\

Trans-Dec (K) & \textcolor{red}{i was in school all the way up to a few years ago and most of us still used non - mechanical ones . (0)} \\

Trans-MLM (K) & i have been in school for a few years , but not really . (0) \\

Trans-AR (K) & how old are you ? i have never been in school for a few years . (0) \\

PF\&FG-free (K) &  how old are you ? i have never had a lot of time in school (2) \\

\hline
\end{tabular}
\caption{\label{tab:redditcase1} Responses generated by baselines and our methods trained on the \textbf{Reddit} dataset(million-scale and 100K). Human evaluation scores are given at the end of each generated reply.}
\end{table*}

\begin{table*}[h]
\small
\begin{tabular}{ll}
\hline
\hline
\textbf{Dialog History} & i have read all of the books , which character do you not like ? [SEP] it 's gon na be the dornish \\
& and greyjoys . and probably brienne . you will see . \\
\textbf{Gold Response} & nah , they have only just started a game of thrones so it must be one of the starks , dan , or tyrion . \\
\hline
SEQ2SEQ-MMI & i do not know why you are being downvoted . (0)\\

Trans-ED & i have read all of the books , and it 's gon na be dornish . (0) \\

Trans-Dec & i have not read the books yet , but i will definitely check them out . (2) \\

Trans-MLM  & i have not read the dornish and greyjoys , but you will see them all as well . (0) \\

Trans-AR  &  brienne is not a fictional character , she 's an actual person . (2) \\

PF\&FG-free & i have read all of the dornish , but that 's just me . (1) \\

\hline
Trans-ED (K) & i am not sure what you are talking about , but it 's just a joke . (0)\\

Trans-Dec (K) & i do n't know , i have read all of the dornish books . (0) \\

Trans-MLM (K) & i do not think it 's a good idea to say that . (0) \\

Trans-AR (K) & i do not think it 's gon na be the best . (0) \\

PF\&FG-free (K) &  i do not know , but it 's a lot of people ? (1) \\

\hline
\end{tabular}
\caption{\label{tab:redditcase2} Responses generated by baselines and our methods trained on the \textbf{Reddit} dataset(million-scale and 100K). Human evaluation scores are given at the end of each generated reply.}
\end{table*}

\begin{table*}[h]
\small
\begin{tabular}{ll}
\hline
\hline
\textbf{Dialog History} & religion , unfortunately . most of the world is still religious . science is opening people 's minds \\
&  but there is still a long way to go . [SEP] religious people can be scientists as well , i do not know \\
& why reddit thinks they are mutually exclusive . \\
\textbf{Gold Response} & nah , they have only just started a game of thrones so it must be one of the starks , dan , or tyrion . \\
\hline
SEQ2SEQ-MMI & i do not know what you are talking about , but it 's a religion . (0)\\

Trans-ED & i do not know why you are being downvoted , but there is a long way to go . (1) \\

Trans-Dec & i do n't know why you think reddit thinks they are mutually exclusive . science is opening people 's \\
& minds , but there still a long way to go (0) \\

Trans-MLM  & because they are not mutually exclusive . it 's a lot of people , but reddit is the only place where science \\
& and religion exist ? (0) \\

Trans-AR  &  religious people can be scientists as well , i do not know why reddit thinks they are mutually exclusive . (0) \\

PF\&FG-free & they are not mutually exclusive . i am just saying that most of the world is still open to science (2) \\

\hline
Trans-ED (K) & i am not sure what you are talking about , but it 's just a joke . (0)\\

Trans-Dec (K) & i do n't know why reddit thinks they are mutually exclusive , but i am sure there is a way . (0) \\

Trans-MLM (K) & i do not know why reddit thinks it 's a science . (0) \\

Trans-AR (K) & they can be scientists , but not all of the world is religious . (1) \\

PF\&FG-free (K) &  i do not know why reddit thinks they are mutually exclusive . science is the only thing that matters to \\
&  you , right ? (2) \\

\hline
\end{tabular}
\caption{\label{tab:redditcase3} Responses generated by baselines and our methods trained on the \textbf{Reddit} dataset(million-scale and 100K). Human evaluation scores are given at the end of each generated reply.}
\end{table*}

\begin{table*}[h]
\small
\begin{tabular}{ll}
\hline
\hline
\textbf{Dialog History} & is that easy to do ? [SEP] usually the ubuntu installer will set it up perfectly for you \\
\textbf{Gold Response} & and then i would select at start-up ? \\
\hline
SEQ2SEQ-MMI & how do i install it ? \\

HRED-MMI & i am not sure what you mean by that \\

Trans-ED & is there a way to do it without installing the ubuntu installer ? i am not sure . \\

Trans-Dec & ok , i will try that . thanks for the help ! :-) it 's working now \\

Trans-MLM  & is there a way to do it with the live cd ? i am not sure if that 's what you are looking for , but how \\
& does one install ubuntu on \\

Trans-AR  &  ok , i will give it a try . thank you very much for your help ! \\

PF\&FG-free & how do i set it up for me ? :) . is there a way to install ubuntu on my laptop with the live cd \\

\hline
Trans-ED (K) & i am not sure what you mean by that , but it 's just a little bit more specific \\

Trans-Dec (K) & ok , i will try that . thanks for your help ! :-) $<$url$>$ \\

Trans-MLM (K) & i am not sure what you mean by that , but it 's a good idea . \\

Trans-AR (K) & i am not sure what you are talking about , but it 's possible that the installer will be able to boot from cd \\

PF\&FG-free (K) &  i am not sure what you mean by setting it up perfectly . is there a way to do that ? \\

\hline
\end{tabular}
\caption{\label{tab:ubuntucase1} Responses generated by baselines and our methods trained on the \textbf{Ubuntu} dataset(million-scale and 100K). }
\end{table*}

\begin{table*}[h]
\small
\begin{tabular}{ll}
\hline
\hline
\textbf{Dialog History} & usually if on the same machine , $<$url$>$ [SEP] : i am getting error . phpmyadmin not in this server \\
\textbf{Gold Response} & you installed it from the software center , or with aptitude or apt-get , or some other way ? \\
\hline
SEQ2SEQ-MMI & what is the error you are getting ? \\

HRED-MMI & i have no idea what you are talking about \\

Trans-ED & i am not sure what you are trying to do , but it 's a good idea . \\

Trans-Dec & did you install phpmyadmin from the repos ? if so , it should be in $<$path$>$ \\

Trans-MLM  & i am not sure what you are trying to do , but it should be in $<$path$>$ . try sudo dpkg - \\
& reconfigure phpmyadmin \\

Trans-AR  &  i am not sure what you are trying to do , but it 's possible . \\

PF\&FG-free & what is the exact error messages , you need to be honest i havenot used it . \\

\hline
Trans-ED (K) & i am not sure what you mean by " phpmyadmin $<$path$>$ \\

Trans-Dec (K) & i am not sure what phpmyadmin is , but it 's probably in $<$path$>$ \\

Trans-MLM (K) & i am not sure what you mean by " phpmyadmin - $>$ $<$path$>$ . $<$url$>$ \\

Trans-AR (K) & i do not know what phpmyadmin is , but it 's in the repos \\

PF\&FG-free (K) &  i am not sure if it 's on the same machine , you can use phpmyadmin \\

\hline
\end{tabular}
\caption{\label{tab:ubuntucase1} Responses generated by baselines and our methods trained on the \textbf{Ubuntu} dataset(million-scale and 100K). }
\end{table*}

\end{document}